\documentclass{article}
\usepackage[utf8]{inputenc}
\usepackage{tablefootnote}
\usepackage{natbib}
\setcitestyle{authoryear}

\usepackage{graphicx}
\usepackage{authblk}

\title{An Introduction to Neural Architecture Search for Convolutional Networks}

\author[1, *]{George Kyriakides}
\author[1]{Konstantinos Margaritis}
\affil[1]{Department of Applied Informatics, University of Macedonia, Thessaloniki 55236, Greece}
\affil[*]{ge.kyriakides@uom.edu.gr}

\date{}

\begin{document}

\maketitle

\begin{abstract}
Neural Architecture Search (NAS) is a research field concerned with utilizing optimization algorithms to design optimal neural network architectures. There are many approaches concerning the architectural search spaces, optimization algorithms, as well as candidate architecture evaluation methods. As the field is growing at a continuously increasing pace, it is difficult for a beginner to discern between major, as well as emerging directions the field has followed. In this work, we provide an introduction to the basic concepts of NAS for convolutional networks, along with the major advances in search spaces, algorithms and evaluation techniques.
\end{abstract}

\section{Introduction}
During the past decade deep learning has proven to be very effective in various automation tasks \citep{jing2019neural, lample2017playing, long2017deep}. This is largely due to a combination of its ability to learn features relevant to each domain, the plethora of data available from various disciplines, as well as the increase of computing power in the form of accelerators, such as GPUs and TPUs. Although able to automatically distinguish features, deep learning demands architectural engineering i.e. design of neural architectures that can be efficiently and effectively trained to perform well in a given task. This process is highly dependent on human analysts, as a solid understanding of deep learning as well as the application domain is needed in to to design the networks. Thus, the next logical step in automating machine learning is to automate the networks' design process.

Neural Architecture Search (NAS) is the research field that emerged from various efforts of automating the architectural design process. NAS has been able to produce many state-of-the-art networks \citep{Zoph2016, Zoph2018}, while advances in the field have proposed methods that do not require vast amount of resources \citep{Pham2018}. A NAS procedure can be divided in several components, each of them contributing to the trajectory, as well as to the result of the search. The most distinct components are the \textit{Search Space}, the \textit{Optimization Method} and the \textit{Candidate Evaluation Method}. 

Search space defines the networks that can be examined to produce the final architecture. This can be argued to be the most important decision when designing NAS procedures, as it can greatly reduce the complexity of the search and thus the computational requirements. Selecting a quality search space can enable even random search to produce highly-performing architectures. Nonetheless, defining such spaces has two caveats. First, it requires prior knowledge about the dataset, which means that it is not suitable for novel domains. Second, it introduces bias, as it excludes various architectures which have not been explored by humans and may offer better performance.

The optimization method dictates how to explore the search space, which can greatly influence the efficiency of the search, as well as the effectiveness of the final proposed architecture. Being an optimization problem, NAS exhibits the classical exploration-exploitation trade-off. Thus choosing an appropriate optimization strategy can ensure that the chosen search space is explored sufficiently, while the proposed architecture is as close to the global optimum as possible.

Finally, the candidate evaluation method is responsible for comparing intermediate results and helping the optimization strategy to choose between various options during the search phase. As evaluating the properties of deep learning architectures can be expensive due to the training required, various methods are utilized in order to speed up the process.

\section{Search Spaces}
In this section we discuss the various search spaces utilized in NAS works. We first present the global search space. Following, we discuss specific cell-based search spaces, as well as

\subsection{Global Search Space}
There are two approaches regarding the NAS search spaces for convolutional layers. The first, more general strategy is to allow the optimization algorithm to generate arbitrary networks. This is referred to as the \textit{Global Search Space} and the search becomes a \textit{macro-architecture search}. Here, for each network the algorithm decides on each layer's type, hyper-parameters, and connections with other layers. In DeepNEAT \citep{miikkulainen2019evolving} the authors explore the space of arbitrarily connected convolutional layer blocks. For each block, they optimize the convolutional layer kernel size and filter number, as well as the dropout rate, the existence of a pooling layer after each convolution, and the weight scaling of each layer. Furthermore, for each architecture they optimize a number of training hyper-parameters, such as the learning rate, momentum, utilization of nesterov accelerated gradient and data augmentation parameters\citep{ruder2016overview, taylor2018improving}. An algorithm called NASH \citep{Elsken2018NASH}, utilizes network morphisms \citep{wei2016network} to generate networks that explore the global search space. Other uses of network morphisms for macro-architecture search can be found in \citep{Cai2018, Jin2019} Another method, DENSER \citep{denser2019} creates arbitrarily connected networks of at least 3 layers, up to a maximum of 30 convolutional or pooling layers and up to 10 fully-connected layers. Finally in \citep{Byla2019}, the authors generate sequentially connected deep networks, consisting of convolutional, pooling and batch normalization layers.

\subsection{Micro or Cell-based Search Space}
\begin{figure}
\centering
\includegraphics[scale=0.6,  angle =90 ]{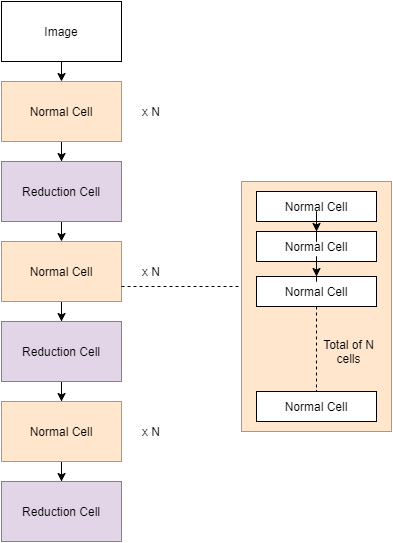}
\caption{The NASNet outer skeleton.}
\label{fig:NASNet skeleton}
\end{figure}

The second approach to search spaces entails the definition of various rules in order to narrow the space. This also limits the optimization algorithm's freedom to generate arbitrary networks. Motivated by repeating patterns in human designs, the basic outer skeleton of the network is fixed and consists of repeated blocks of layers, called \textit{Cells} (Figure \ref{fig:NASNet skeleton}). A common cell-based search space is the NASNet search space, proposed in \citep{Zoph2018}. NASNet consists of two types of cells, Normal and Reduction cells. In such spaces, the algorithms search for cell architectures, as cells are just repeated. Thus, instead of designing a very big network, two smaller architectural patterns are created (one for normal and one for reduction cells) and the network is compiled by replacing each cell placeholder with the proposed cell. In \citep{Zoph2018}, a reduction block follows every N normal cells and each cell receives the outputs of the two previous cells as inputs. Other works utilize variations of this space, such as \citep{Zhong2018} which uses N normal cells followed by a max pooling operation, while \citep{Liu2018} utilizes N normal cells with stride 1 (where applicable, such as convolutional or pooling layers) followed by a cell with stride 2. Another example of a NASNet-like search space can be found in \citep{dong2018dpp}, where N cells are followed by an average pooling layer with stride 2 and dense connections between the layers. Another approach entails the utilization of highly performing human-crafted architectures as outer skeletons \citep{cai2018path}.

\subsection{Hierarchical Search Spaces}
The main advantage of cell-based search spaces is that they are able to leverage prior domain knowledge and thus produce better performing architectures, as the search becomes more efficient \citep{Pham2018}. Their main disadvantage is that bias is introduced, and possibly even better performing architectures are omitted from the search. Furthermore, in some domains where other networks characteristics are important such as latency, it has been shown that global search is better suited for the task \citep{tan2019mnasnet}. In this work, the authors utilize a hierarchical approach, where the skeleton consists of various non-repeating cells. Similar approaches of hierarchically building the networks are followed by two other works. The first utilizes N levels of design patterns called motiffs \citep{Liu2018hierarchical}. Each pattern consists of a directed acyclic graph. In level $i$, the graph's nodes are motiffs of level $i-1$. Level 2 motiffs utilize various types of layers as their nodes (such as convolution and pooling layers). Note that the paper's experimental section is concerned with searching for cells only. The second work employs blueprints of skeletons and cells, which are jointly optimized \citep{miikkulainen2019evolving}. It can be seen as a special case of the previous work, with 3-level motiffs.

\section{Optimization Methods}
Optimizing the structure of neural networks, as well as their weights has been a research topic long before the advent of deep learning \citep{stanley2002evolving}. Recent works utilize a number of techniques in order to design the networks' architecture, as weights are usually trained by various optimizers. In this section we discuss these techniques and how they can be applied to NAS.

\subsection{Evolutionary Algorithms}
\begin{figure}
\centering
\includegraphics[scale=0.6]{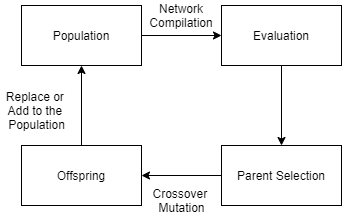}
\caption{The procedure at each evolutionary generation.}
\label{fig:evolutionaries}
\end{figure}

\textit{Evolutionary Algorithms} utilize a population of individuals to create offspring of increasingly better performance. Individuals, just like real organisms are defined by their genes and genomes. Each gene contains a piece of information about the problem being optimized. Genes can mutate in order to alter their information and individuals can reproduce by crossover, thus creating offspring with genes from both parents. In NAS, genes usually contain information about each layer in the network (layer type and parameters), as well as the connections between the layers. The main difference between the various implementations lie in their crossover and mutation operators. Moreover, their choice of gene and genome representations greatly influence the available operations for mutation and crossover. Mutation-only methods are the most popular amongst researchers, as they do not demand the rigorous logging of gene origins. Furthermore, mutations can be implemented with function-preserving transformations, such as network morphisms, allowing the inheritance of trained weights from parents to offspring. Finally, as evolutionary algorithms usually have a pool of offspring to evaluate at each generation, they are the most straight-forward family of optimization methods to parallelize, being almost "embarrassingly parallel". This allows the speedup and the scale up of the algorithms, enabling the exploration of greater regions in the search space. The basic procedure that every evolutionary method follows is depicted in Figure \ref{fig:evolutionaries}.

\citet{miikkulainen2019evolving} employ genetic algorithms and co-evolution of species \citep{moriarty1997forming} in order to evolve outer skeleton blueprints, as well as species of cells (referred to as modules). This work utilizes mutation, by adding or removing connections and layers, as well as by peturbing a layer's parameters. Crossover is implemented by marking a genome's history, thus allowing the alignment of homologous genes. At each generation, the worst-performing individuals are removed, while the rest of the population reproduces and is replaced by its offspring. A similar philosophy of designing simultaneously the outer skeleton, as well as the repeating patterns within it is followed in \citep{Liu2018hierarchical}, although only the mutation operator is used. Here, the authors propose a hierarchical design of many levels, as opposed to the two levels proposed in \citep{miikkulainen2019evolving}. Instead of replacing the old population, this work continuously expands it, while utilizing tournament selection (tournament size equal to 5\% of the population) in order to determine each offspring's parents. In \cite{Real2019} the authors propose a regularized evolution to design convolutional cells. Again, only the mutation operator is utilized, with a tournament selecting the parents of each offspring. Here the authors replace the oldest individual with a single offspring at each cycle. They argue that this forces architectures that achieve consistently high performance to survive, while filtering out those acheiving good performance once, due to luck or initial conditions. \citet{kyriakides2020regularized} utilize a combination of regularized evolution from \citet{Real2019} and genome representations from \citet{miikkulainen2019evolving} in order to conduct global search. The work of \citet{Elsken2018NASH} propose another evolutionary scheme utilizing only mutation. The current best model is selected as a parent a number of children are created, by mutating the parent through network morphisms.

\subsection{Reinforcement Learning}
\begin{figure}
\centering
\includegraphics[scale=0.55 ]{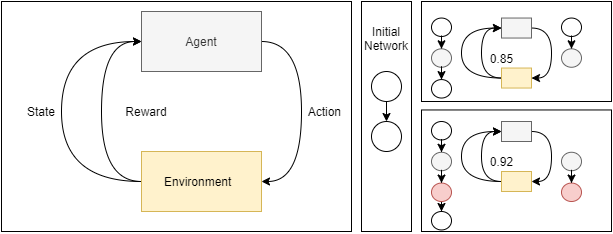}
\caption{Expanding a simple network with reinforcement learning.}
\label{fig:reinforcement}
\end{figure}

\textit{Reinforcement learning} (RL) is a sub-field of machine learning where instead of mapping inputs to a target, the algorithms attempt to map inputs to optimal actions. An agent (either physical or virtual) interacts with its environment by selecting actions, in an effort to maximize its reward. The agent may be rewarded after each action, or only at specific points. By repeatedly interacting with the environment and observing its rewards, the agent improves its ability to select highly-rewarding actions or series of actions (policy). Reinforcement learning methods are formulated as Markov Decision Processes, where the set of states $S$, the set of actions $A$, and the reward discount factor are known. In NAS, reinforcement learning has been successfully employed as a method to design architectures. Their main differences (except from the RL algorithm utilized) lie in the definition of the state and action spaces. Their reward is usually a function of the estimated performance of the generated network. Action spaces consist of layer type, layer parameter selection, and inter-layer connections, with the search space dictating and usually being identical to the action space. The state can be seen as the current generated architecture, even if it is not yet complete (in cases where the architecture is sequentially sampled, i.e. layer by layer). Figure \ref{fig:reinforcement} shows a simple visualization of how the agent may expand a small network with only an input and output layer, by adding two additional layers. One drawback of reinforcement learning when compared to evolutionary methods is their less straight-forward parallelization.

One of the first papers to use reinforcement learning, and the one that coined the NAS acronym is the work of \citet{Zoph2016}. Here, the authors utilize a Recurrent Neural Network (RNN) controller in order to sequentially generate architectures, using an autoregressive approach. The controller is trained with the REINFORCE algorithm. At each step, the controller selects the filter height and width, the stride height and width, the number of filters and previous layers in order to generate skip-connections. In a later work \citep{Zoph2018}, the authors propose and search in the NASNet search space. Here, the controller predicts the two inputs and two operations (one for each input) as well as a method to combine the output of the operations. As mentioned previously, the action space is dictated by the search space. There are a total of 12 discrete operations available to choose from, as well as the identity. The controller is trained with the Proximal Policy Optimization (PPO) algorithm. \citep{Cai2018} also utilize REINFORCE to train a bidirectional Long Short-Term Memory (Bi-LSTM) RNN, with a simple two-actions action space; widening or deepening each layer. Both actions are implemented by function-preserving network morphisms. Their work is further expanded in \citep{cai2018path}, where branching between layers is also part of the action space. \citet{tan2019mnasnet} also employs the same RNN as in \citep{Zoph2018} to sequentially sample architectural parameters, while trying to optimize both the latency, as well as the accuracy of the generated networks on mobile devices. \citet{Real2019} directly compare their work with \citet{Zoph2018}, showing that both RL and evolutionary algorithms can produce competent architectures. The evolutionary algorithm exhibits better any-time performance and requires less time to run, while the RL approach exhibits a smaller amount of variance in the performance of the final architecture.

\subsection{Bayesian Optimization}
In \textit{Bayesian Optimization} a predictive model of the objective function is utilized in order to choose the most promising arguments and then evaluate them on the actual function. In the context of NAS, the objective function is usually the network's performance and the arguments are its architecture. \citet{kandasamy2018neural} propose a novel distance metric to compute similarity between networks and thus can employ Gaussian Process-based Bayesian Optimization, by using the metric as a kernel function. The distance metric can be efficiently computed via an optimal transport program. In this work, the authors employ an evolutionary algorithm in order to optimize the acquisition function and then evaluate the optimal points (neural architectures) on the original objective function, namely they train and evaluate the produced architectures. The results of the evaluations are used to update the predictive model. Another approach utilizes Sequential Model-Based Optimization in order to explore the NASNet search space \citep{Liu2018}. In this work, the authors search for architectures of increasing complexity (increasing the number of layer blocks in a cell) while utilizing a surrogate model to predict the compiled network's accuracy, given the encoding of the cell's architecture. At each step, a new set of candidate architectures is generated in line with the work of \citet{Zoph2018}. The architectures' performance is predicted with the surrogate model and the top $K$ are selected in order to be evaluated by training and testing on the dataset. \citet{Luo2018} utilize an encoder-decoder scheme in order to generate architctures, while employing a surrogate model in order to predict the performance of a candidate, utilizing the encoder's embeddings as inputs. By maximizing the predicted performance, the authors utilize the decoder in order to translate the embeddings to a neural architecture. Finally, \citet{zela2018towards} utilize a combination of Bayesian Optimization and Hyperband, a multi-arm bandit strategy for hyper-parameter optimization, in order to jointly optimize the neural architecture and its hyper-parameters. In order to more efficiently utilize computational resources, when evaluating the candidates more promising candidates are allocated more resources while less promising candidates are discarded early in the evaluation process.

\subsection{One-Shot Methods}
\begin{figure}
\centering
\includegraphics[scale=0.48]{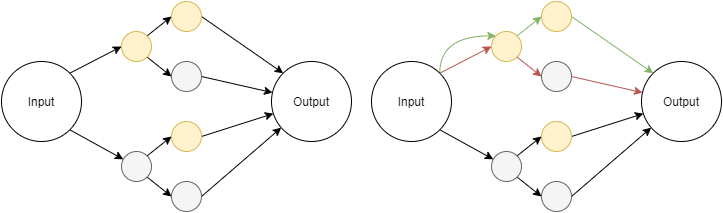}
\caption{Small One-Shot hyper-graph example (left). Sampling two architectures sharing the same weights on one layer (right). The two architectures (red and green arrows) share the weights on the first yellow layer.}
\label{fig:oneshot}
\end{figure}

\textit{One-Shot Methods} try to leverage the fact that weights of a neural architecture are mostly feature maps and can be utilized in many different but closely-related domains, as shown by transfer learning \citep{Zoph2018}. Thus, they usually employ a hyper-graph which includes all possible architectures defined by the search space and individual architectures are sampled by various methods. The benefit of these methods is that weights can be shared amongst different architectures, thus requiring considerably less resources in order to conduct the search. A visual example is seen in Figure \ref{fig:oneshot}. The cost is somewhat greater than training a single architecture. \citet{Pham2018} utilize Reinforcement Learning in order to sample paths within the hyper-graph of NASNet, while alternating between updating the hyper-graph's weights and the controller's weights. Compared to the original work of \citet{Zoph2016}, this paper acheives a speedup of over 1000. Instead of utilizing a controller to generate the architectural samples, \citet{bender2019understanding} sample architectures at random, in an effort to better understand one-shot methods. They conclude that they are effective even with simple gradient descent and do not require complex optimization methods, such as reinforcement learning. An approach utilizing gradient descent is presented in DARTS \citep{Liu2018darts}. Instead of considering individual paths in the hyper-graph, all paths are utilized simultaneously and weights are assigned to each path. These path-weights are optimized concurrently with the traditional layer-weights by gradient descent. After the training concludes, the paths with the lowest weights are pruned in order to generate the final architecture. Although this method extends the classic gradient-descent based approach to other network parameters besides layer weights and is thus conceptually simple, it requires that the whole hyper-graph's weights are kept in memory at any time during search. Thus, it exhibits the classical space-time trade-off. The work of \citet{xie2018snas} further expands DARTS, by encoding each layer level as a one-hot vector and sampling one vector for each level. Thus, a Monte-Carlo estimate of the gradient is computed, while requiring that only the active layer weights are kept on-memory. This greatly reduces the memory requirements of the algorithm, compared to \citet{Liu2018hierarchical}.

\subsection{Meta-Learning Hyper-Networks}
In an effort to further reduce the need to train individual architectures, some researchers propose to train \textit{Meta-Learning Hyper-Networks} that directly generate sufficient weights for the candidate architectures \citep{Brock2017}. Here, instead of candidates being paths of a hyper-graph, they are trained by a hyper-network, which can greatly reduce the memory required. As such, this method can also be considered part of the One-Shot family. \citet{zhang2018graph} further expand the idea, utilizing graph neural networks and hyper-networks. First, a neural architecture is sampled and a graph is derived from the architecture. Following, a graph neural network (GNN) is created, homomorphic to the sampled architecture's graph. Each node in the graph is a RNN which passes messages along its edges and stores an internal embedding vector, which is updated recurrently. Using these embeddings, the hyper-network generates the weights for the original sampled architecture. Note that the hyper-network is a simple Multi-Layer Perceptron.

\section{Candidate Evaluation Methods}

\textit{Candidate Evaluation Methods} aim to evaluate candidate architectures, in order to allow optimization methods to choose between them. As NAS is usually concerned with finding the best performing architecture, the most straight-forward approach is to train and evaluate each candidate on the selected dataset. Nonetheless, this is a computationally expensive process and demands many GPUs to run in parallel for many days \citep{Zoph2016, Zoph2018, Real2019}. Furthermore, during the search phase, the absolute performance of each candidate is note very useful. Instead, the relative performance of the various options is of paramount importance. This has lead to the experimentation with alternative methods in order to evaluate the quality of an architecture. In this section we present the most well-known methods that have been utilized in order to speed up the process of evaluating various architectures.

The most intuitive way to reduce the computational cost of evaluating the relative performance of various architectures is to train them for a smaller number of epochs. This has been used in various works with success, as they were able to generate architectures that outperformed the state-of-the-art \citep{cai2018proxylessnas, liang2018evolutionary, Real2019, Liu2018, zela2018towards, miikkulainen2019evolving}. Although the method produces good empirical results, some have argued that it should not be used when the search and final training epochs differ drastically \citep{zela2018towards}. In \citep{kyriakides2020effect}, the authors show that given a sufficiently small discrepancy between the training epochs during the search phase and the final architecture training, there is a positive correlation in relative rankings between architectures. Hyper-networks \citep{Brock2017, zhang2018graph} and network morphisms \citep{Elsken2018NASH, Cai2018, Jin2019, Cai2018, cai2018path} can be utilized to provide a warm-start to candidate architectures. In turn, this reduces the number of epochs required to train the networks to convergence.

The second most intuitive way to reduce the cost of evaluating an architecture's quality is to alter the training data. As in the case of transfer learning, many researchers choose to conduct search in a less computationally demanding dataset and then transfer their results to the original \citep{Zoph2018} with success. Furthermore, in cell search spaces many researchers reduce the number of cells stacked in each block and the number of filters where applicable (model augmentation) \citep{Real2019, Liu2018hierarchical, Liu2018}.

Another approach is to utilize predictive models, in order to predict the performance of candidate architectures and use those predictions to guide the search. This has been discussed previously in Section 3.3, where Bayesian Optimization methods utilize surrogate models. Nonetheless, actual evaluations are needed in order to train the models \citep{kandasamy2018neural, Liu2018, Zoph2018, zela2018towards, Luo2018}. Furthermore, the utilization of surrogate models has the potential to transfer knowledge about good architectures between datasets. 

Finally, as shown in \citet{bender2019understanding}, weight-sharing amongst candidate architectures has been demonstrated to provide positive ranking correlation to architectures trained from scartch. This has been leveraged by the one-shot and weight-sharing family of methods, where either a hyper-graph is trained with gradient descent \cite{Liu2018darts} or individual paths are chosen and the corresponding weights are updated \citep{Pham2018, xie2018snas, bender2019understanding}. Concluding the algorithm presentation sections, Table \ref{comparisons_table} contains an overview of the algorithms presented in this work. Table \ref{abbrev_table} contains an explanation of the abbreviations used.

\begin{table}
\caption{Overview of various algorithms.}\label{comparisons_table}
\begin{tabular}{p{3.5cm}rrr}
Paper & Search Space & Optimization & Evaluation Speed Up \\
\hline
\cite{bender2019understanding} &Cell &O-S RS &MA, RE, WS \\
\cite{Brock2017} &Cell &O-S HN RS &RE, TL, WG \\
\cite{Byla2019} &Global &ACO &RE \\
\cite{Cai2018} &Global &RL &RE, TL \\
\cite{cai2018path} &Cell &RL &MA, RE, TL \\
\cite{cai2018proxylessnas} &Global &O-S GD &RE, WS \\
\cite{denser2019} &Global &EA &RE, TL \\
\cite{dong2018dpp} &Cell &BO &MA, RE, SM, TL \\
\cite{Elsken2018NASH} &Global &EA &RE, WS \\
\cite{Jin2019} &Global &BO &SM \\
\cite{kandasamy2018neural} &Global &BO &RE, SM \\
\cite{kyriakides2020regularized} &Global &EA &MA, RE \\
\cite{liang2018evolutionary} &Global &EA &RE, WS \\
\cite{Liu2018} &Cell &BO &SM, RE \\
\cite{Liu2018darts} &Cell &O-S GD &MA, RE, TL \\
\cite{Liu2018hierarchical} &Hierarchical &EA &MA, RE, TL \\
\cite{Luo2018} &Cell &BO &MA, RE, TL, WS \\
\cite{miikkulainen2019evolving} &Hierarchical &EA &RE \\
\cite{Pham2018} &Global \& Cell &RL &RE, WS \\
\cite{Real2019} &Cell &EA &MA, RE, TL \\
\cite{tan2019mnasnet} &Hierarchical &RL &MA, RE, TL \\
\cite{xie2018snas} &Cell &O-S GD &MA, RE, WS, TL \\
\cite{zela2018towards} &Cell &BO &SM, RE \\
\cite{zhang2018graph} &Cell &HN RS &MA, RE, WG \\
\cite{Zhong2018} &Cell &RL &MA, RE, TL \\
\cite{Zoph2016} &Global &RL &MA, RE \\
\cite{Zoph2018} &Cell &RL &MA, RE, TL \\

\end{tabular}
\end{table}

\begin{table}
\caption{Abbreviations}\label{abbrev_table}
\begin{tabular}{l r | l r}
Abbreviation &Meaning &Abbreviation &Meaning \\
\hline
ACO &Ant Colony Optimization \tablefootnote{\cite{dorigo2006ant}} &MA &Model Augmentation \\
BO &Bayesian Optimization &RE &Reduced Epochs \\
EA &Evolutionary Algorithms &SM &Surrogate Model \\
GD &Gradient Descent &TL &Transfer Learning \\
HN &Hyper-Networks &WG &Weight Generation \\
O-S &One-Shot &WS &Weight Sharing \\
RL &Reinforcement Learning & & \\
RS &Random Search & & \\

\end{tabular}
\end{table}

\section{Comparing Algorithms}
Due to the nature of NAS and Deep Learning, it is difficult to directly compare various algorithms published in different papers. As results are usually reported in terms of performance, which depends not only on the NAS algorithm but also on various other factors, comparisons can be unjust. The most frequently variable aspects are the dataset pre-processing pipeline, the implementation, and training hyper-parameters of the networks, as well as the search space. A recent publication by \citet{lindauer2019best} proposes a checklist of best practices, concerning three major categories. First, releasing code for the algorithm, the training pipeline, and the search space, as well as hyper-parameters and random seeds. Second, reporting important details about the experiments, such as hyper-parameter tuning, required times, and experimental setup. Finally, when comparing different NAS methods, all methods should be evaluated by utilizing the same dataset, search space, training code, and hyper-parameters. Furthermore, the authors propose to control for confounding factors, as well as to run ablation studies, compare performance over time, compare to random search, perform multiple experimental runs and use tabular or surrogate benchmarks for in-depth evaluations.

\subsection{Benchmarks}

As mentioned earlier, it is difficult to directly compare results of various algorithm implementations. NAS benchmarks attempt to establish a standardized environment, where different algorithms can be evaluated fairly and quickly. These benchmarks consist of pre-computed network evaluations for a complete search space. This allows researchers without access to expensive compute accelerators to conduct research on NAS, as each generated architecture's performance can be retrieved through a table look-up. This greatly reduces computational requirements, as well as wall-clock time for NAS experiments.

The first cell search NAS benchmark was published by \citet{ying2019bench}, named NAS-Bench-101. It contains 423,000 unique architectures. Each cell consists of a maximum of 7 layers and 9 edges (connections between layers). Each layer can be either a 3x3 convolution, a 1x1 convolution or a 3x3 max-pooling operation. The models are trained and tested on the CIFAR-10 dataset \citep{krizhevsky2009learning}. Each model is trained for 4, 12, 36, and 108 epochs, three times for each epoch schedule. This provides researchers with enough data to simulate re-sampling of the same architecture by the NAS algorithms. For each architecture the training, validation, and testing accuracy are reported after the full and half training epochs, as well as the training time and number of trainable parameters. 

An extension to the original benchmark named NAS-Bench-201 was proposed by \citet{dong2020bench}, introducing 5 layer types (skip-connect, zeroize, 1x1 convolution, 3x3 convolution, and 5x5 average-pooling). Cells consist of 4 layers, with no restrictions on the maximum number of edges, resulting in 15,625 distinct networks. Each network is evaluated on both CIFAR-10 and CIFAR-100 \citep{krizhevsky2009learning}, as well as on ImageNet-16-120 \citep{chrabaszcz2017downsampled}. Furthermore, the benchmark contains more details than the original NAS-Bench-101, reporting model latency and FLOPs, as well as loss and accuracy after every training epoch. An interesting advantage is that this benchmark can be utilized by weight-sharing NAS algorithms.

Finally, a benchmark dedicated to one-shot methods is proposed in \citep{zela2020bench} named NAS-Bench-1shot1. Instead of evaluating new architectures, the authors propose a mapping between models generated through one-shot methods and the models already evaluated in the original NAS-Bench-101. By utilizing choice blocks \citep{bender2019understanding}, the authors are able to define different search spaces, based on the number of parents each choice block has. There are a total of three different search spaces mapped to the original benchmark, resulting in 6,240, 29,160, and 363,648 unique models. As the benchmark is an extension of NAS-Bench-101, the same metrics are reported for each architecture.

\section{Conclusions}

In this work we have presented the most well-known works in the field of Neural Architecture Search (NAS) and presented their basic components; their search spaces, their optimization methods, and their candidate evaluation methods. As the field is relatively new, many of these components are expected to evolve, change, or become obsolete. Furthermore, there is a certain overlap and interdependence between them. Search spaces greatly influence the way architectures are represented by optimization methods and may introduce bias. Representations in turn, influence how much freedom optimization methods have to explore the search space. Evaluation methods may further introduce bias in the search, as they can augment the fitness landscape of the search space. Furthermore, certain evaluation methods may influence the choice of optimization methods. This is the case with weight-sharing hyper-graphs where gradient descent is able to perform well, rendering the use of more sophisticated optimization methods obsolete. Benchamrking datasets, although greatly beneficial in fairly and quickly comparing NAS algorithms, may also induce bias in the form of overfitting. Following best-practices and releasing both code as well as implementation details when publishing NAS methodologies can ensure an ethical approach to the subject, although it does not guarantee bias-free methods. Finally, works cited here are concerned with finding good architectures about contemporary convolutional neural networks. Given that the building blocks of neural networks (layers, activation functions, regularizations) are also evolving, it is expected that NAS methods will also adapt to these changes.

Although NAS aims to simplify and automate the process of designing neural networks, it seems that instead of removing the decision process of architectural design completely, it replaces it with another. Programming languages evolved from machine language, to assembly and high-level languages while shifting software design decisions to higher levels. It is probable that neural networks will also follow a similar path, shifting decisions from layer and parameter selection to search space, optimization and evaluation methods, to even higher hierarchical levels.

\bibliography{references}
\end{document}